 \RequirePackage{amsthm}
\documentclass[sn-mathphys,pdflatex,Numbered]{sn-jnl}

\usepackage{float}%
\usepackage{graphicx}%
\usepackage{multirow}%
\usepackage{amsmath,amssymb,amsfonts}%
\usepackage{amsthm}%
\usepackage{mathrsfs}%
\usepackage[title]{appendix}%
\usepackage{xcolor}%
\usepackage{textcomp}%
\usepackage{manyfoot}%
\usepackage{booktabs}%
\usepackage{algorithm}%
\usepackage{algorithmicx}%
\usepackage{algpseudocode}%
\usepackage{listings}%
\usepackage{subfigure}%

\usepackage{xurl}

\theoremstyle{thmstyleone}%
\begin{document}

\title[Graph Analysis Using a GPU-based Parallel Algorithm: Quantum Clustering]{Graph Analysis Using a GPU-based Parallel Algorithm: Quantum Clustering}


\author[1]{\fnm{Zhe} \sur{Wang}}

\author[1]{\fnm{Zhijie} \sur{He}}

\author*[1]{\fnm{Ding} \sur{Liu}}\email{liuding@tiangong.edu.cn}

\affil[1]{\orgdiv{Department of Computer Science and Technology}, \orgname{Tiangong University}, \postcode{300387}, \state{Tianjin}, \country{China}}




\abstract{The article introduces a new method for applying Quantum Clustering to graph structures. Quantum Clustering (QC) is a density-based unsupervised learning method that determines cluster centers by constructing a potential function. In this method, we develop the Graph Gradient Descent algorithm to find the centers of clusters for graph analysis. GPU parallelization is utilized for computing potential values. We also conducted comparative experiments on five widely used datasets and evaluated them using four indicators. The results show superior performance of our method. Finally, we discuss the influence of the crucial parameter $\sigma$ on the experimental results.}




\maketitle

\section{Introduction}
\label{intro}
Graph Clustering, also known as network clustering, is a technique for partitioning a graph into clusters or communities of nodes based on their structural properties. Graph clustering is used in various applications such as social network analysis\citep{bu2018gleam}, image segmentation\citep{li2021image,jia2020fast}, bioinformatics\citep{smirnov2021magus}, and more. The goal of graph clustering is to group the nodes in a way to maximizes the similarity within the group and minimizes the similarity between them. These two similarities are usually measured using various metrics such as Modularity\citep{newman2004finding}, Normalized Mutual Information(NMI)\citep{strehl2002cluster}, Adjusted Rand Index(ARI)\citep{hubert1985comparing} and Fowlkes\-Mallows Index(FMI)\citep{fowlkes1983method}. A diverse range of algorithms exists for graph clustering, encompassing techniques such as K-Means \citep{arthur2007k}, Spectral Clustering \citep{shi2000normalized, von2007tutorial, knyazev2001toward}, DBSCAN \citep{ schubert2017dbscan}, Louvain \citep{blondel2008fast, dugue2015directed}, Label Propagation Algorithm (LPA) for localized community detection \citep{raghavan2007near}, BIRCH \citep{zhang1997birch}, AGENS \citep{zhang2013agglomerative, fernandez2008solving}, and more recently proposed Deep Graph Clustering methods like AGC \citep{zhang2019attributed} and GCC \citep{fettal2022efficient}. The challenging problem in graph clustering is that we need to cluster its basic structures and use these structures for clustering purposes, which need more efficient clustering algorithms. However, 
QC is a very effective clustering algorithm to uncover subtle changes in the underlying data. In a recent study, quantum clustering has been used to predict the health status of lithium-ion batteries with special degradation paths\citep{gao2022quantum}, as well as digital neutron-gamma discrimination using quantum clustering\citep{lotfi2019neutron}. Additionally, quantum clustering has also been applied in the field of biology\citep{sequeira2019human, gottlieb2022time}.

The Quantum clustering\citep{horn2001algorithm} is a quantum-inspired clustering method based on the Schr\"{o}dinger equation. QC calculates the so-called potential function to reveal the structure of the data. While the potential function depends entirely on the parameter $\sigma$ and we discuss the $\sigma$ in section \ref{sec:Discussion}. QC has been extensively demonstrated and experimented in our previous work\citep{liu2016analyzing}, and show its superior performance. In order to find the minimum node in the graph structure, we develop a so-called Graph Gradient Descent(GGD) algorithm and we will describe the algorithm in detail in the section \ref{sec:Algorithm}. In section \ref{sec:Application}, we benchmark the performance of QC on five datasets and compare it with eight other algorithms. All implementation code is available in the \citep{Wangzhe}

\section{Related works}
\label{sec:Related works}
\begin{figure}[ht]
  \centering
    \includegraphics[width=5.2in,height=2.88in]{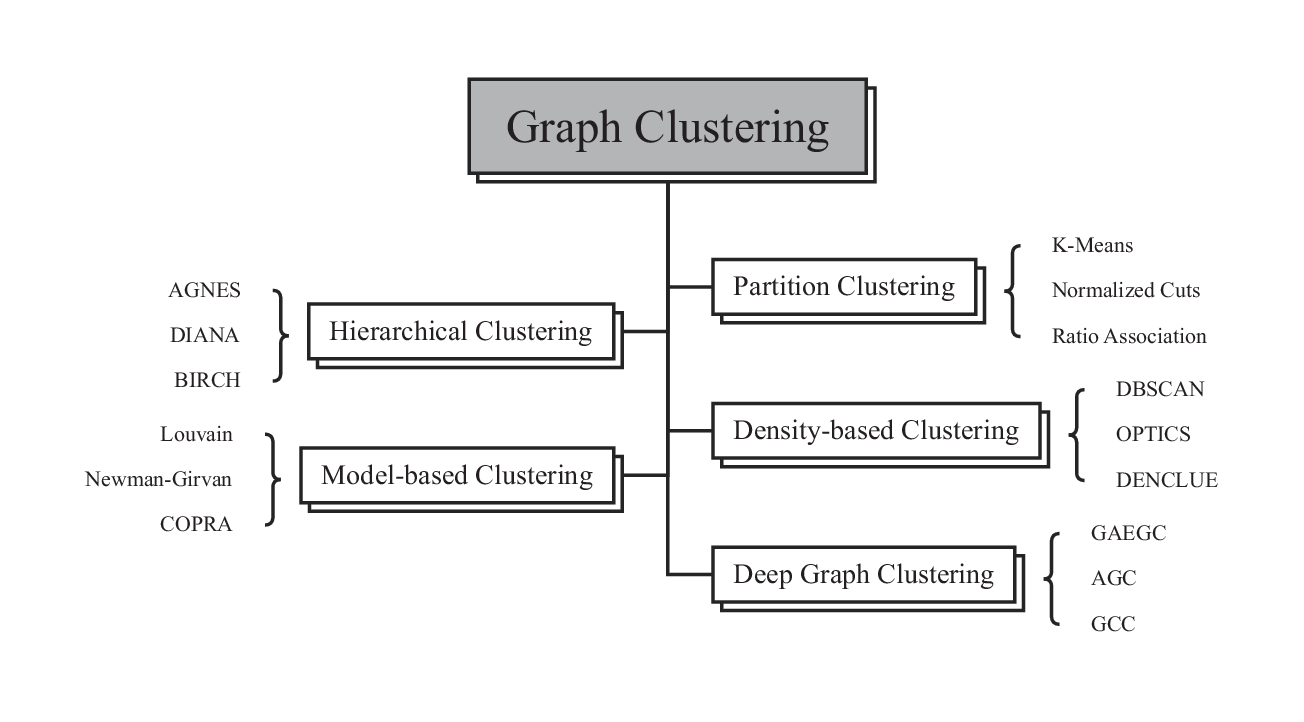}
    \centering
  \caption{\centering Overview of Graph Clustering;}
  \label{fig:1} 
\end{figure}

The graph clustering algorithm can be mainly divided into five categories. i.e, Partition Clustering, Hierarchical Clustering, Density-based Clustering, Model-based Clustering and Deep Graph Clustering. These methods all has its unique advantages and application scenarios. Representative algorithms for each clustering method are shown in Fig.\ref{fig:1}. And QC can be regarded as a density-based clustering method. 

\subsection{Partition Clustering}

Partition Clustering divides the graph into multiple subgraphs, and each subgraph contains nodes belonging to the same class. This method is usually implemented using spectral clustering methods, such as spectral clustering based on the Laplacian matrix. Spectral Clustering can effectively handle clusters with non-convex and irregular shapes and is robust in the presence of noisy data, making it highly practical. in practice. In lately work \citep{berahmand2022novel} use a new version of the spectral cluster, named Attributed Spectral Clustering (ASC), ASC use the Topological and Attribute Random Walk Affinity Matrix (TARWAM) as a new affinity matrix to calculate the similarity between nodes.

\subsection{Hierarchical Clustering}
Hierarchical Clustering is a strategy of cluster analysis to create a hierarchical of clusters. HC first builds a binary tree and node information is stored in each node. The algorithm starts from such a leaf node, gradually traverses towards the root node, and classifies similar nodes into one category. Also the algorithm can traverse from the root node to the leaf nodes. This divides HC into two categories, agglomerative (bottom-up) and divisive (top-down)\citep{li2022ensemble}. Ref.\citep{dogan2022k} propose a novel linkage method, named k-centroid link. 

\subsection{Density-based Clustering}
In \citep{kriegel2011density} Density-based Clustering defined as a non-parametric approach where clusters are considered as high density regions of density function. The steps of density-based clustering is to find the core point\citep{braune2015density}, and then divide the nodes in the adjacent area into a cluster and assign the border point to the cluster where its adjacent core point is located. Finally, remove noise points. \citep{kriegel2011density} Imagine the density-based clusters as the set of points resulting from "cutting" the probability density function of the data at some density level.

\subsection{Model-based Clustering}
This method models the graph clustering problem as a probabilistic model and use methods such as EM algorithm and Bayesian inference to learn the model parameters and obtain the clustering results. This method is usually implemented using models such as Gaussian mixture models and latent Dirichlet allocation\citep{mcnicholas2016model}.

\subsection{Deep Graph Clustering}
In \citep{liusurvey}, Deep Graph Clustering is introduced as a method that utilizes neural networks to encode the nodes of a graph. The encoding process involves transforming the graph's node attributes and adjacency matrix into meaningful embeddings, leveraging both the structural and attribute information. Subsequently, a clustering method is employed to partition the encoded nodes into distinct, disjoint clusters. The objective is to group nodes based on their similarities and capture underlying patterns within the graph.

\section{Method}
\label{sec:method}

\subsection{Algorithm}
\label{sec:Algorithm}
Quantum Clustering\citep{liu2016analyzing,nasios2007kernel,horn2001algorithm} is a new machine learning algorithm based on the Schr\"{o}dinger equation. In our work, we choose the time-independent  Schr\"{o}dinger equation Eq(\ref{equ:1}) \citep{feynman1965feynman}. We use this equation to explore graph structures at a deeper level. The algorithm process can be decomposed into the following steps. 

\begin{equation}
\label{equ:1}
H\psi\left(x \right)=\left(-\frac{\hbar^{2}}{2m}\nabla^{2} +v(x)\right)\psi \left(x \right)
                    =E\psi \left(x \right)
\end{equation}

\noindent
Here H denotes Hamiltonian operator, which is an operator that describes the energy of a quantum system. \emph{$\psi(x)$} denotes Wave function, which is the fundamental physical quantity that describes a quantum system. and \emph{v(x)} denotes potential function, which is describing the probability density function of the input data\citep{nasios2007kernel}. Given a Gaussian wave function Eq(\ref{equ:2}), use the Schr\"{o}dinger equation to calculate the potential function. Here $\sigma$ denotes the width parameter. In the graph structure, we assume that when there is an edge between two nodes, the distance between the two nodes is the weight of the edge. If there is no edge between two nodes, then the distance between them is a maximum value higher than other weights.

\begin{equation}
\label{equ:2}
\psi \left(x\right)=\sum_{i}e^{-\left(x-{x}_{i} \right)^{2}/2\sigma ^{2}}
\end{equation}

Thus, the potential function \emph{v(x)} could be solved as:

\begin{equation}
\label{equ:7}
\begin{aligned}
v(x)&=E+\frac{\sum_{i}(e^\frac{-(x-x_{i})^2}{2\sigma^2} \cdot\frac{(x-x_{i})^2}{2\sigma^2}-e^\frac{-(x-x_{i})^2}{2\sigma^2}\cdot\frac{1}{2})}{\sum_{i}e^\frac{-(x-x_{i})^2}{2\sigma^2}}\\
&=E-\frac{1}{2}+\frac{1}{2\sigma^2\psi(x)}\sum_{i}(x-{x}_{i})^2e^\frac{-(x-x_{i})^2}{2\sigma^2}\\
&\approx \frac{1}{2\sigma^2\psi(x)}\sum_{i}(x-{x}_{i})^2e^\frac{-(x-x_{i})^2}{2\sigma^2}
\end{aligned}
\end{equation}

\begin{algorithm}
\label{Alg:1}
\caption{Calculating the Potential function for each data point}
    \begin{algorithmic}
        \Require $graph:$ graph structure represented by adjacency matrix which $graph(i,j)$
        
        \par \hspace{4.8em} represents the weight between the "i-th" and "j-th" nodes,
        
            \par \hspace{3.5em} $\sigma:$ width parameter,
            \par \hspace{3.5em} $n:$ the number of data points,
            \par \hspace{3.5em} $i:$ index of graph node
        \Function {POTENTIAL}{$i$}
             \State $sum1 \gets 0$;
             \State $sum2 \gets 0$;
               \For{$j = 1 \to n$}
                    \State $dist \gets graph(i, j)$;
                    \State $sum1 \gets sum1+dist^2\cdot e^{-dist^2/2\sigma^2}$;
                    \State $sum2 \gets sum2+e^{-dist^2/2\sigma^2}$;
               \EndFor
               \State $v(i) \gets \frac{1}{2\sigma^2}\cdot \frac{sum1}{sum2}$;
               \State \Return {$v(i)$};
            \EndFunction

\end{algorithmic}
\end{algorithm}

\begin{algorithm}
\label{Alg:2}
\caption {Graph Gradient Decent algorithm}
\begin{algorithmic}
    \Require $v(0..n)$: the potential values for all nodes in the graph structure
    \Function{GRAPH\_GRADIENT\_DECENT}{$v(0..n-1)$} 
        \For{$i = 0 \to n$}
            \State $neighbor$ $\gets$ the collection of nodes adjacent to node i;
            \State $low\_potential\_value\_index$ $\gets$ the index of the node with the minimum 
            \State potential function value within the $neighbor$ array;
            \For{$j = 1 \to len(neighbor)$}
                \If{$v(neighbor(j)) < low\_potential\_value$}
                    \State $low\_potential\_value\_index \gets j$;
                \EndIf
            \EndFor
            \State $results(i) \gets low\_potential\_value\_index$;
        \EndFor
        \For{$i = 0 \to n$}
                \State $results(i) \gets$ $FIND\_CLUSTER\_CENTER(i,results(0..n-1))$;
        \EndFor
        \State\Return{$results(0..n-1)$}
    \EndFunction
\end{algorithmic}
\end{algorithm}

\begin{algorithm}
\label{Alg:3}
\caption {Find Cluster Center algorithm}
\begin{algorithmic}
    \Require $i$: the indices of nodes in the graph structure,
    \par \hspace{1.5em} $results(0..n-1)$: intermediate results in Algorithm 2
    \Function{FIND\_CLUSTER\_CENTER}{i, results$(0..n-1)$}
        \If{$i != results[i]$}
            \State $results[i] \gets $ $FIND\_CLUSTER\_CENTER(results[i],results(0..n-1))$;
        \EndIf
        \State\Return{$i$};
    \EndFunction
\end{algorithmic}
\end{algorithm}

\begin{figure}
    \includegraphics[width=5in,height=2in]{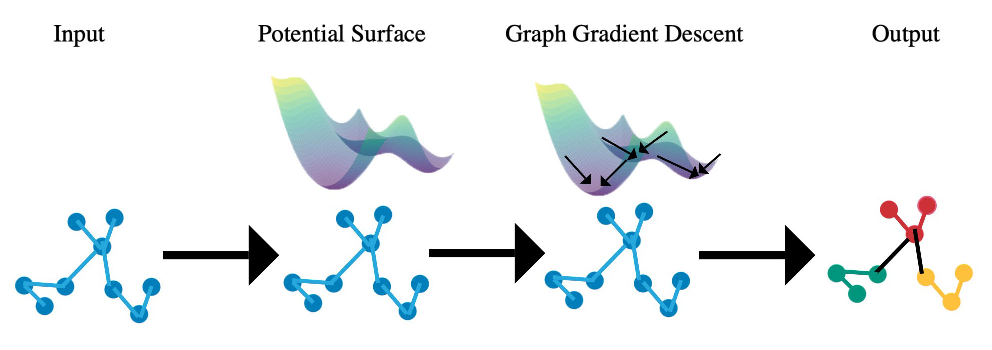}
  \caption{\centering Schematic diagram of the GGD Algorithm;}
  \label{fig:2} 
\end{figure}

\begin{algorithm}
\label{Alg:4}
\caption{Quantum clustering for graph analysis}
    \begin{algorithmic}
        \Require $graph:$ graph structure represented by adjacency matrix, 
        \par \hspace{3.5em} $n:$ the number of data points,
        \par \hspace{3.5em} $\sigma:$ {initial parameter}
            \For{$i = 1 \to n$}
                \State $v(i) \gets$ $POTENTIAL(i)$;
            \EndFor              
            \State $labels \gets GRAPH\_GRADIENT\_ DECENT(v(0..n-1))$;
            \State\Return{$labels$};
\end{algorithmic}
\end{algorithm}
In our study, to tackle the issue of locating local minima on graph structures, a task unattainable through conventional gradient descent methods, we introduce a novel optimization algorithm specifically crafted for the exploration of local optimal solutions. In this algorithm, we have devised a gradient descent path for each node within the graph structure to guide them towards nodes with the lowest potential energy. Initially, each node forms an independent cluster, and the central node exhibits a lower potential energy than its surrounding nodes. This fundamental principle aligns with the effectiveness of quantum clustering in analyzing the graph structure. Subsequently, each node iterates through its neighboring nodes to identify the node with the lowest potential energy. If the potential energy of a node is lower than that of the initial node, the initial node is assigned to the cluster where the target node exists. The GGD algorithm's schematic diagram is shown in Fig. \ref{fig:2}. 
In Fig. \ref{fig:2}, we illustrated the process of the algorithm using a simple artificial dataset, which includes inputs, outputs, and the potential function surfaces constructed by the algorithm. Additionally, we depicted the principle of gradient descent algorithm for finding local optima on these potential function surfaces. The pseudocode for this portion is presented in Algorithm 2, and the time complexity of Algorithm 2 is depend on the density of the graph structure. 

Algorithm 1 and 4 provide the pseudocode for computing the potential function and outline the fundamental framework of the entire algorithm. The time complexity of Algorithm 1 is $O(n)$. For Algorithm 2, we start by obtaining an array $neighbor$ containing all adjacent nodes' indices to node $i$. Then, through iterating over the $neighbor$ array, we identify the index of the node adjacent to node $i$ in the graph structure with the minimum potential function value, then storing it in the $result$ array. Lastly, we apply the algorithm of logical search tree in Algorithm 3 to locate the indices of nodes corresponding to local minima. Logical search tree organizes the relationships between nodes using an array consisting of node indices. Each node is directly affiliated to its clustering center. This approach optimizes the height of the logical search tree structure, resulting in a time complexity of $O(h)$ for implementing the code of the logical tree structure, which is approximate to $O(log(n))$. Here, \textit{h} represents the height of the logical search tree, and \textit{n} represents the size of the dataset. For entire GGD algorithm, in most instances, iterating over the neighbors of a single node takes $O(1)$ time. However, in the worst-case scenario where each node is fully connected to all other nodes, the time complexity for iterating the neighbors becomes $O(n)$. Therefore, GGD has a worst time complexity of $O(n^2)$.

\subsection{Parallelized by GPU}
\begin{figure}
  \centering

    \includegraphics[width=3.5in,height=3.0in]{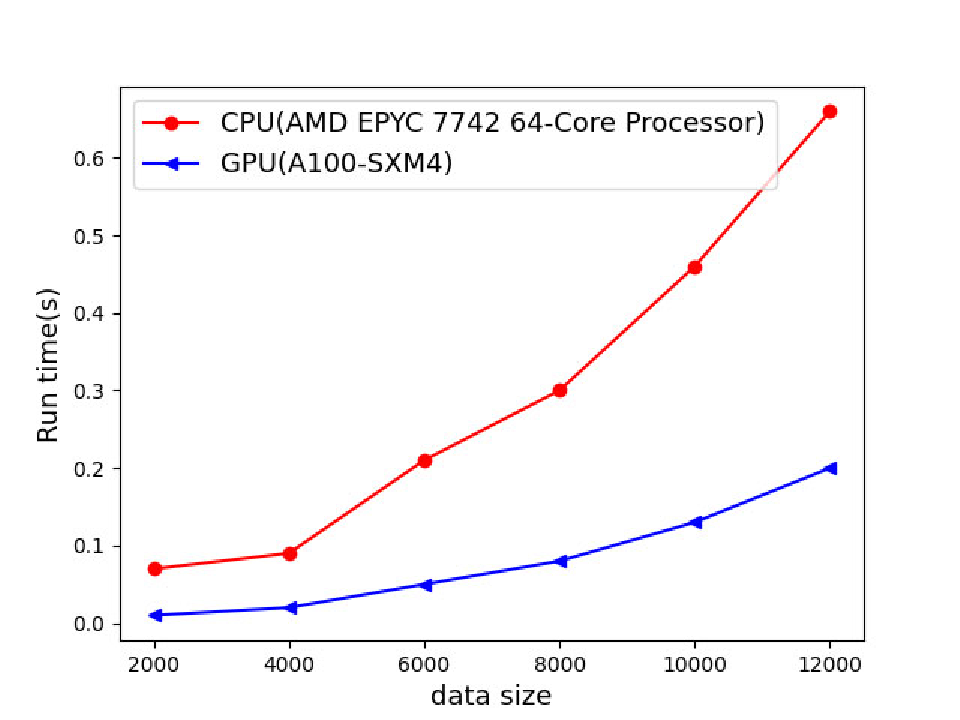}
    \centering
  \caption{\centering Comparison of GPU and CPU Acceleration, through experiments, we found that as the size of data increases, the time taken by the algorithm to compute the potential function on the GPU is significantly lower than the time on the CPU. This demonstrates the notable acceleration effect brought about by computing the potential function on the GPU;}
  \label{fig:3} 
\end{figure}

The most important part of QC algorithm is to calculate the potential value of each data point. So it is very suitable to use GPU for parallelization. In this part, we design experiments to prove its acceleration effect. the GPU version we used for this experiment is A100-SXM4, And the counterpart of CPU version is AMD EPYC 7742 64-Core Processor. We use a series of artificial dataset with different data volumes to complete the experiment. Comparison of GPU and CPU acceleration on a fixed fully-connected graph structure dataset by increasing the number of nodes Fig. \ref{fig:3}. All implementation code is available in the \citep{Wangzhe}.

\section{Application}
\label{sec:Application}
\subsection{Dataset}
\begin{figure}
    \centering
  \subfigure[]{
  
      \label{fig:4:a} 
      \includegraphics[width=2.0in,height=2.0in]{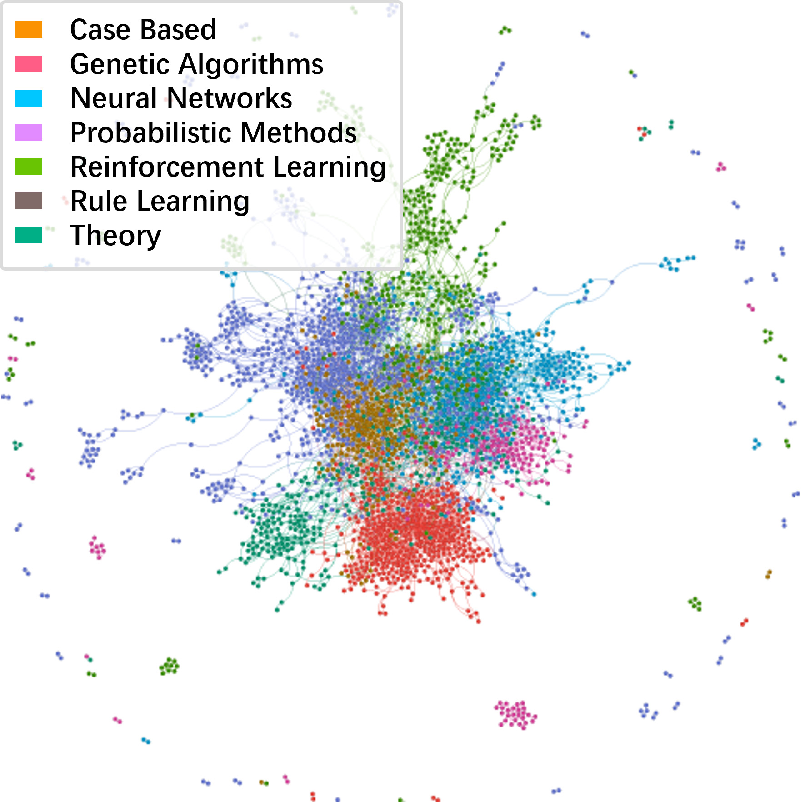}}
    \subfigure[]{
      \label{fig:4:b} 
      \includegraphics[width=2.0in,height=2.0in]{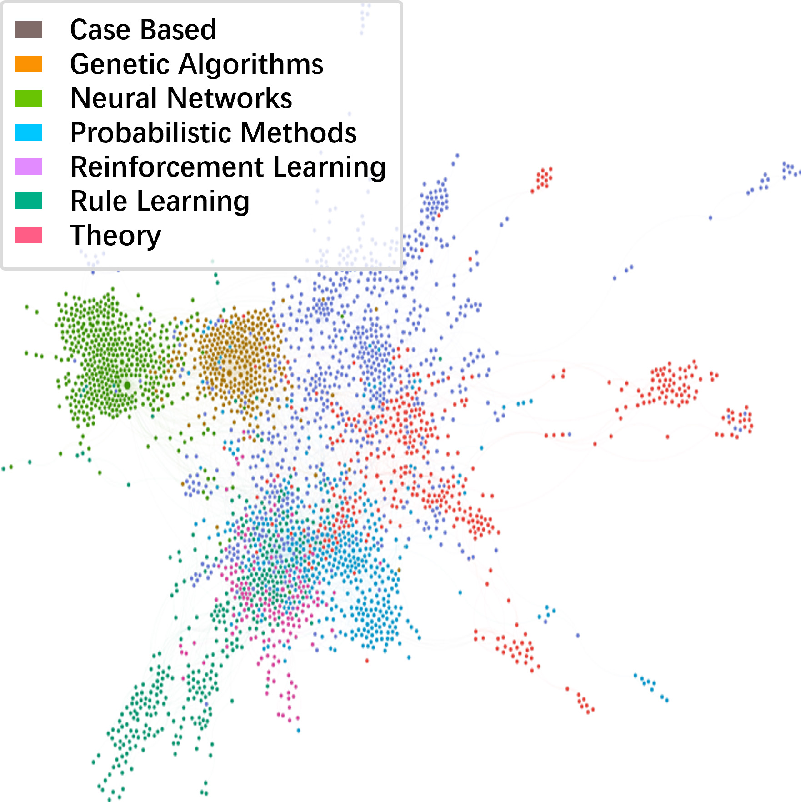}}
      \subfigure[]{
      \label{fig:4:c} 
      \includegraphics[width=2.0in,height=2.0in]{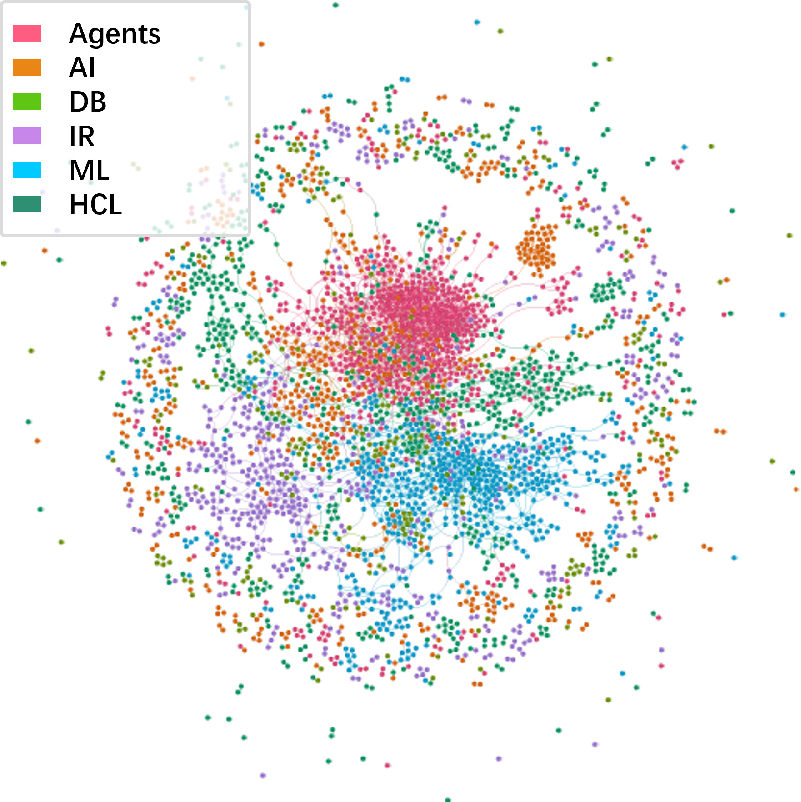}}
      \subfigure[]{
      \label{fig:4:d} 
      \includegraphics[width=2.0in,height=2.0in]{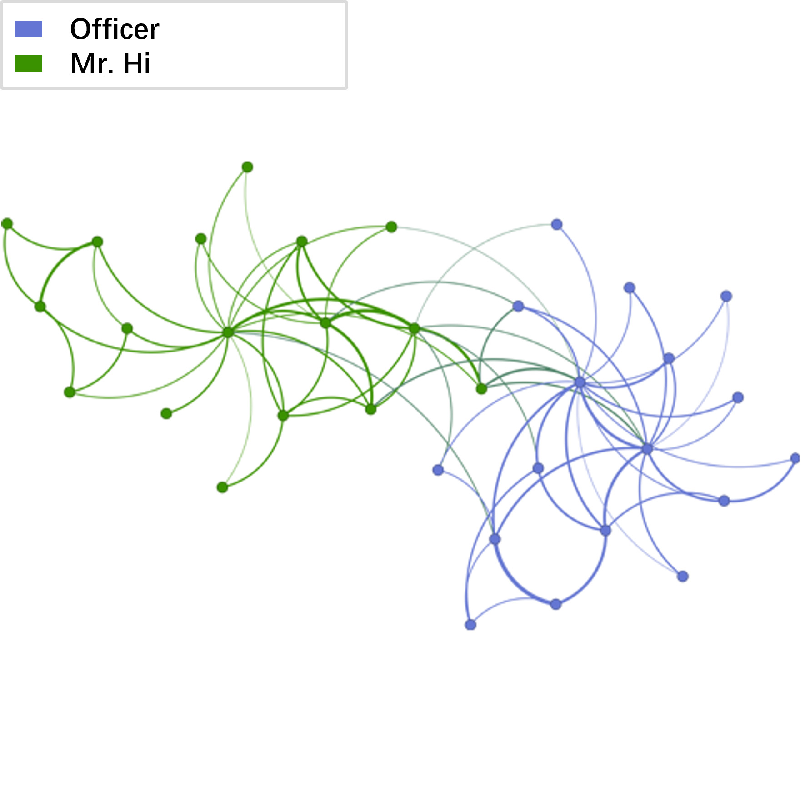}}
      \subfigure[]{
      \label{fig:4:e} 
      \includegraphics[width=2.0in,height=2.0in]{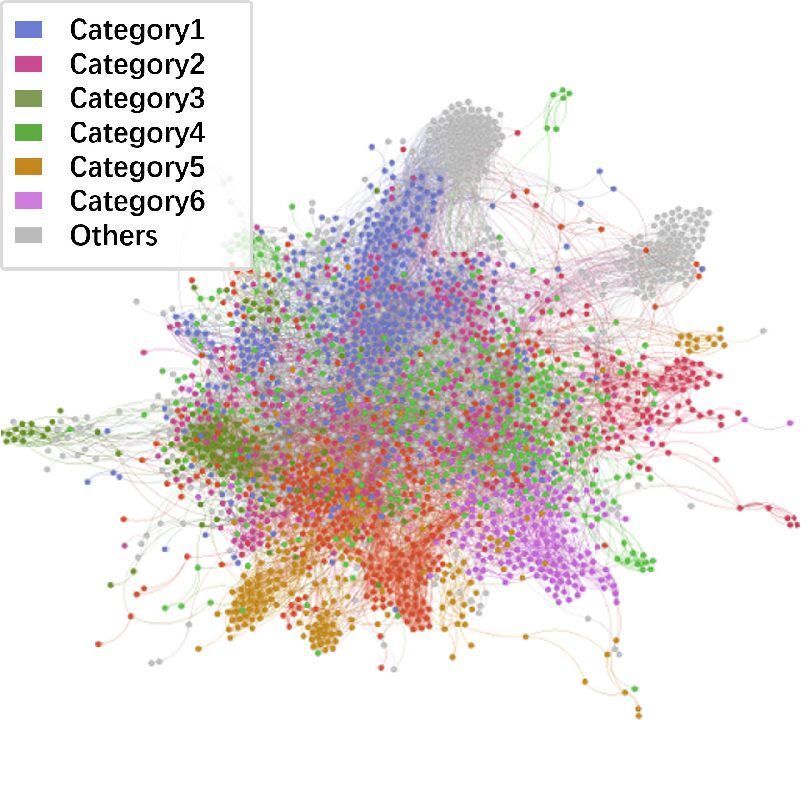}}
      
    \caption{The visualization of datasets for our experiment. ForceAtlas2 is used as layout demonstration algorithm. Different colors represent different clusters. (a) Cora dataset; (b) Cora-ML dataset; (c) Citeseer dataset; (d) Karate Club datset; (e) Wiki dataset; Note that since the Wiki dataset lacks actual class labels, we have opted to represent the classes in the legend using categories such as "category1", "category2", and so on;}
    \label{fig:4} 
  \end{figure}

To evaluate the proposed method, we choose five widely-used datasets in this experiment, i.e., Cora\citep{sen2008collective}, Citeseer\citep{sen2008collective}, Karate Club\citep{zachary1977information}, Cora-ML, Wiki. To provide a more intuitive presentation and analysis of these datasets, 
we use the Gephi , a leading open-source graph exploration toolkit. For visualization purposes, we employ ForceAtlas2\citep{jacomy2014forceatlas2} as the layout demonstration algorithm. It is a force-directed layout that not only strikes a better balance between performance and quality but also reveals a clearer structures of graphs. The results are presented in Fig. \ref{fig:4}, where each color represents a distinct class.

\subsubsection{Cora \& Cora-ML \& Citeseer datasets}
The Cora dataset is kind of citation network of 2708 scientific publications and 5278 citation relationships covering important topics in the field of computer science, including 7 classes, i.e., machine learning, artificial intelligence, databases, networks, information retrieval, linguistics, and interdisciplinary fields. Each node represents a paper, and the edges between nodes represent citation relationships. 

The Cora-ML dataset is a variant of the Cora dataset, which is a citation network containing 2995 scientific publications and 8158 citation relationships where each node represents a paper and edges represent citation relationships. The difference between Cora-ML and Cora is that Cora-ML also includes category labels for papers in the machine learning field.

Similar to the Cora dataset, Citeseer dataset consisting of 3327 papers and 4676 citation relationships downloaded from the Citeseer digital library which classified into 6 classes.
\subsubsection{Karate Club dataset}
Karate Club dataset represent a social network consisting of 34 nodes and 78 edges. The social network was observed and recorded by Zachary in 1977, and represents the relationships between members of a Karate Club. This social network is commonly used as a benchmark dataset. In this social network, each node represents a member of the karate Club, and the edges represent social connections between members. According to Zachary's records, the social network eventually split into two communities.
\subsubsection{Wiki dataset}
The Wiki dataset containing 2405 Wikipedia pages and 12761 link relationships and these pages can divided into 17 classes. Each page is represented as a node, and the link relationships between nodes form the graph.
\subsection{Evaluation method}
We employ four metrics to evaluate the performance of QC: Modularity\citep{newman2004finding}, Adjusted Rand Index(ARI)\citep{hubert1985comparing}, FowlkesMallows Index(FMI)\citep{fowlkes1983method}, and Normalized Mutual Information(NMI)\citep{strehl2002cluster}. These 4 metrics can be divided into 2 classes, called internal and external measures. While the formers are often used when there are no real label, and latters are often used when there are real label. Next, we introduce these indicators in detail. 
\subsubsection{Modularity}
Modularity reflects the degree of connection between nodes\citep{newman2004finding}. A desirable cluster partition should demonstrate strong connections within clusters while minimizing connections between clusters. In this scenario, the value of Modularity would be significantly high, indicating that the quality of community division is better.

The formula for Modularity is as follows:

\begin{equation}
    Q = \dfrac{1}{w} \sum_{i,j}\left(A_{ij} - \gamma \dfrac{w_iw_j}{w}\right)\delta_{c_i,c_j}
\end{equation}

\noindent where A is the adjacency of network, $w_i$ represents the degree of a node, $w$ is the total weight, $\delta$ represents the Kronecker symbol and $\gamma$ is the resolution parameter.
\subsubsection{ARI}
ARI is a commonly used external evaluation metric in cluster analysis. The interval of ARI value range from -1 to 1 \citep{hubert1985comparing}. Where number -1 indicates complete disagreement between the clustering results, number 0 indicates the clustering result is classified randomly, and number 1 indicates complete agreement between the clustering results and the real classification.

The formula for ARI is as follows:

\begin{equation}
    ARI = \frac{RI - Expected\_RI}{max(RI) - Expected\_RI}
\end{equation}

\noindent RI is the Rand Index, $Expected\_RI$ is the expected value of the Rand Index under the null hypothesis of random clustering. The term $(max(RI) - Expected\_RI)$ represent a normalization factor.
\subsubsection{FMI}
FMI is a measure of the similarity between a clustering result and the real class labels\citep{fowlkes1983method}. It is defined as the geometric mean value of the precision and recall between the clustering result and the real class labels.

The FMI is calculated as:

\begin{equation}
    FMI = \frac{TP}{\sqrt{(TP + FP) * (TP + FN)}}
\end{equation}\par
 \
\noindent where TP is the number of true positive, FP represent the number of false positive, FN is the number of false negative.

\subsubsection{NMI}
NMI is a normalization of the Mutual Information (MI) to scale the results between 0 (no mutual information) and 1 (perfect correlation)
The NMI is calculated as follow\citep{strehl2002cluster}:

\begin{equation}
    NMI(labels\_true, labels\_pred) = \frac{MI(labels\_true, labels\_pred)}{\sqrt{H(labels\_true) \cdot H(labels\_pred)}}
\end{equation}

\noindent where $H$ represents Entropy, MI is Mutual Information. And $MI(labels\_true, labels\_pred)$ represents Mutual Information between two sets of labels and can be calculated with following formula:
\begin{equation}
    MI(labels\_true, labels\_pred) = \sum_{i=1}^{n}\sum_{j=1}^{m}P(i, j)\log{\frac{P(i, j)}{P(i)P(j)}}
\end{equation}

\noindent where $P(i, j)$ denotes the proportion of samples that have a real label of $i$ and a predicted label of $j$ out of the total number of samples. $H(labels\_true)$ and $H(labels\_pred)$ represent the entropies of the real labels and predicted labels, respectively, and can be calculated as follows:

\begin{equation}
    H(labels) = -\sum_{i=1}^{n}P(i)\log{P(i)}
\end{equation}

\noindent where $P(i)$ represents the proportion of samples that have a label of $i$ out of the total number of samples.

\begin{table}[htbp]
\caption{\centering Comparison of QC with other algorithms. The graph adjacency matrix is the graph input, and features refer to node characteristics. With only graph input, GCC's Modularity, NMI, and ARI are 0;}
\label{tab:1}
\begin{tabular}{llccccc}
\hline
Dataset      & Algorithms             & Input               & Modularity     & NMI            & ARI            & FMI  \\ \hline
Cora         & \textbf{QC}            & Graph               & \textbf{0.634} & \textbf{0.401} & \textbf{0.166} & \textbf{0.285}       \\
             & kmeans                 & Graph               & 0.017          & 0.023          & 0.004          & 0.422                \\
             & Louvain                & Graph               & 0.812          & 0.443          & 0.236          & 0.358                \\
             & LPA                    & Graph               & 0.747          & 0.389          & 0.155          & 0.267                \\
             & Spectral Clustering    & Graph               & 0.009          & 0.010          & -0.006         & 0.412                \\
             & AGNES                  & Graph               & -0.001         & 0.001          & 0.000          & 0.423                \\
             & BIRCH                  & Graph               & -0.001         & 0.377          & 0.001          & 0.021                \\
             & AGC                & Graph               & -4.720         & 0.004          & -1e-4          & 0.422                \\
             & GCC                & Graph               & 0.034          & 0.004          & 0.003          & 0.392                \\
             & AGC                & Graph \& Node Feature      & 0.736          & 0.535          & 0.447          & 0.545                \\
             & GCC                & Graph \& Node Feature      & 0.724          & 0.587          & 0.502          & 0.595                \\
Citeseer     & \textbf{QC}            & Graph               & \textbf{0.704} & \textbf{0.343} & \textbf{0.073} & \textbf{0.179}        \\
             & kmeans                 & Graph                  & 0.008          & 0.006          & 0.000          & 0.421                 \\
             & Louvain                & Graph                  & 0.891          & 0.332          & 0.101          & 0.216                \\
             & LPA                    & Graph                  & 0.834          & 0.333          & 0.075          & 0.180                 \\
             & Spectral Clustering    & Graph                  & 0.145          & 0.019          & 0.006          & 0.359                 \\
             & AGNES                  & Graph                  & 0.163          & 0.061          & 0.002          & 0.406                 \\
             & BIRCH                  & Graph                  & 0.015          & 0.352          & 0.001          & 0.022                 \\
             & AGC                & Graph                  & 0.006          & 0.002          & -3e-4          & 0.172                 \\
             & GCC                & Graph                  & 0.011          & 7.4e-4         & 8.9e-4         & 0.407                \\
             & AGC                & Graph \& Node Feature         & 0.610          & 0.339          & 0.266          & 0.426                 \\
             & GCC                & Graph \& Node Feature         & 0.738          & 0.451          & 0.455          & 0.554                 \\
Wiki         & \textbf{QC}            & Graph                  & \textbf{0.361} & \textbf{0.133} & \textbf{0.029} & \textbf{0.205}        \\
             & kmeans                 & Graph                  & 0.049          & 0.029          & 0.003          & 0.314                 \\
             & Louvain                & Graph                  & 0.701          & 0.362          & 0.165          & 0.240                 \\
             & LPA                    & Graph                  & 0.308          & 0.193          & 0.026          & 0.301                 \\
             & Spectral Clustering    & Graph                  & 0.114          & 0.141          & 0.021          & 0.320                 \\
             & AGNES                  & Graph                  & 0.049          & 0.048          & 0.006          & 0.316                 \\
             & BIRCH                  & Graph                  & 0.049          & 0.483          & 0.000          & 0.009                 \\
             & AGC                & Graph                  & 0.107          & 0.020          & 2.6e-5         & 0.075                 \\
             & GCC                & Graph                  & 0.006          & 0.003          & -4.6e-4        & 0.309                 \\
             & AGC                & Graph \& Node Feature         & 0.680          & 0.440          & 0.139          & 0.278                 \\
             & GCC                & Graph \& Node Feature         & 0.688          & 0.548          & 0.329          & 0.392                 \\
Cora\_ML     & \textbf{QC}            & Graph                  & \textbf{0.620} & \textbf{0.405} & \textbf{0.219} & \textbf{0.327}        \\
             & kmeans                 & Graph                  & 0.008          & 0.005          & -0.002         & 0.411                 \\
             & Louvain                & Graph                  & 0.770          & 0.479          & 0.312          & 0.419                 \\
             & LPA                    & Graph                  & 0.718          & 0.421          & 0.206          & 0.322                 \\
             & Spectral Clustering    & Graph                  & 0.014          & 0.017          & -0.002         & 0.404                 \\
             & AGNES                  & Graph                  & 0.001          & 0.001          & 0.000          & 0.415                 \\
             & BIRCH                  & Graph                  & -0.001         & 0.379          & 0.002          & 0.035                 \\
             & AGC                & Graph                  & 1e-4           & 0.004          & 3.4e-4         & 0.414                  \\
             & GCC                & Graph                  & 0.034          & 0.002          & -4.1e-4        & 0.355                \\
             & AGC                & Graph \& Node Feature         & 0.668          & 0.560          & 0.457          & 0.563                 \\ 
             & GCC                & Graph \& Node Feature         & 0.686          & 0.574          & 0.481          & 0.573                 \\
Karate Club & \textbf{QC}             & Graph                  & \textbf{0.334} & \textbf{0.649} & \textbf{0.668} & \textbf{0.832}        \\
             & kmeans                 & Graph                  & -0.013         & 0.093          & 0.007          & 0.658                 \\
             & Louvain                & Graph                  & 0.445          & 0.588          & 0.465          & 0.677                 \\
             & LPA                    & Graph                  & 0.305          & 0.544          & 0.504          & 0.717                 \\
             & Spectral Clustering    & Graph                  & 0.357          & 0.469          & 0.283          & 0.528                 \\
             & AGNES                  & Graph                  & 0.225          & 0.244          & 0.109          & 0.630                 \\
             & BIRCH                  & Graph                  & -0.051         & 0.335          & 0.008          & 0.086                 \\ 
             & AGC                & Graph                  & 0.003          & 0.120          & -0.015         & 0.477                \\ 
             & GCC                & Graph                  & 0              & 0              &  0             & 0.534                \\
             & AGC                & Graph \& Node Feature         & 0.411          & 0.752          & 0.712          & 0.793                 \\ 
             & GCC                & Graph \& Node Feature         & 0.373          & 0.707          & 0.583          & 0.721                \\  \hline
\end{tabular}
\end{table}

\subsection{Performance comparison}
To assess the practical applicability of the QC algorithm, we conducted a comparative evaluation against eight other widely employed graph clustering algorithms. The experimental results are presented in Table. \ref{tab:1}. It can be observed that, in the Cora, Citeseer, Wiki, and Cora\_ML datasets, the performance of the Louvain algorithm slightly surpasses QC, as evident from the provided Table. \ref{tab:1}. The Louvain algorithm was originally proposed by Belgian astrophysicist Vincent Blondel and his colleagues in 2008 \citep{blondel2008fast}. The algorithm uses a greedy algorithm based on modularity optimization, which can quickly detect community structure in large networks. And improved in the paper\citep{dugue2015directed}. Additionally, In \citep{raghavan2007near}, the LPA algorithm was proposed, with performance similar to that of QC. In recent years, when node features are incorporated as additional inputs, both AGC and GCC have demonstrated a little better performance compared to QC. AGC and GCC are graph clustering algorithms that effectively utilize additional information to achieve improved outcomes. AGC focuses on incorporating node features into the clustering process, while GCC emphasizes the utilization of graph structure and node attributes for more accurate clustering. Both algorithms have been shown to produce better results when compared to QC in scenarios where node features are considered as part of the input, although they may require significantly longer computation time. However, in the context of the Karate Club dataset, a noteworthy observation is that three out of four metrics clearly exhibit the superiority of QC over the other algorithms when considering only the Graph as input. The Louvain algorithm surpasses QC only when considering Modularity, but the number of clusters produced by QC in the Karate Club dataset aligns with the real labels. So we can calculate the F1 value, Accuracy and Recall. The F1 value is 0.91, Recall rate is 1 and Accuracy rate is 0.91. Shows a great advantage. 

\begin{figure}
  \centering

    \includegraphics[width=5.0in,height=3.0in]{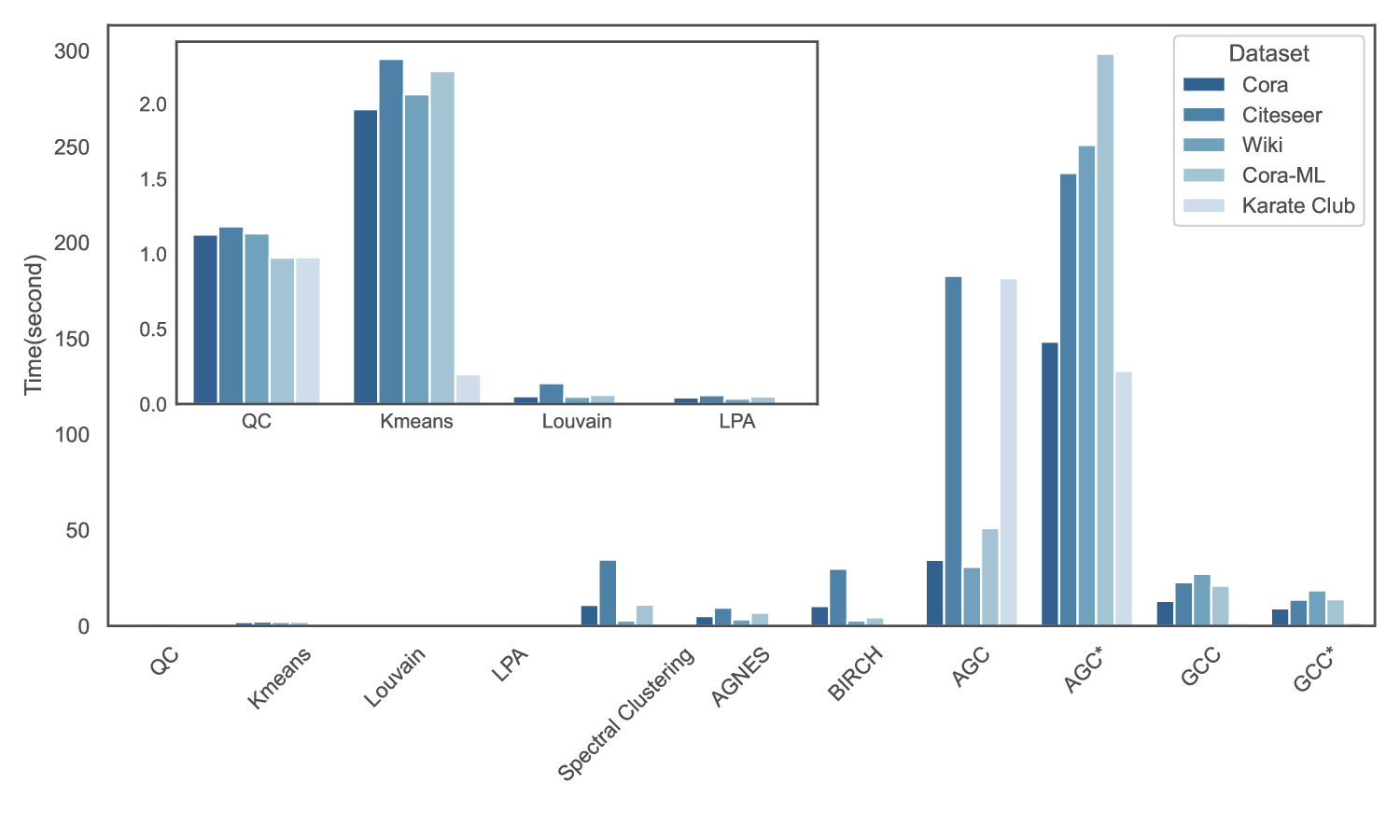}
    \centering
  \caption{\centering Comparison of time consumption among algorithms: It should be noted that the GPU version of QC is represented here. Given that certain algorithms and their corresponding data points appear nearly indistinct in the figure, we have included a zoomed-in subplot on the left for a more precise comparison of time consumption. The star symbol in the upper right corner of these two algorithms, AGC* and GCC*, indicates that they share the same input as the classical algorithms, namely, the adjacency matrix, without requiring the passage of node features;}
  \label{fig:6} 
\end{figure}

In addition, we compared the time consumption of the GPU version of QC with other algorithms in Fig. \ref{fig:6}. Due to the relatively brief time taken by QC, Kmeans, Louvain, and LPA, their time consumption is nearly invisible in the figure. To better compare their performance at this level, we have included a zoomed-in subplot on the left, which provides a magnified view of the relevant data. This zoomed-in representation allows for a more precise assessment of the algorithms' time consumption characteristics. As shown in the figure, QC takes little more time than Louvain and LPA. For Spectral Clustering, AGNES, BIRCH, AGC, and GCC, the time they take varies significantly among different datasets. However, QC and Kmeans demonstrate relatively consistent performance across different datasets in terms of time consumption. Moreover, QC runs faster than Spectral Clustering, AGNES, BIRCH, AGC, and GCC. In all, QC algorithm has relatively good clustering performance, and its time consumption is acceptable.

\section{Discussion}
\label{sec:Discussion}

\begin{figure}[htbp]
  \centering
\subfigure[]{
    \label{fig:5:a} 
    \includegraphics[width=2.1in,height=2.1in]{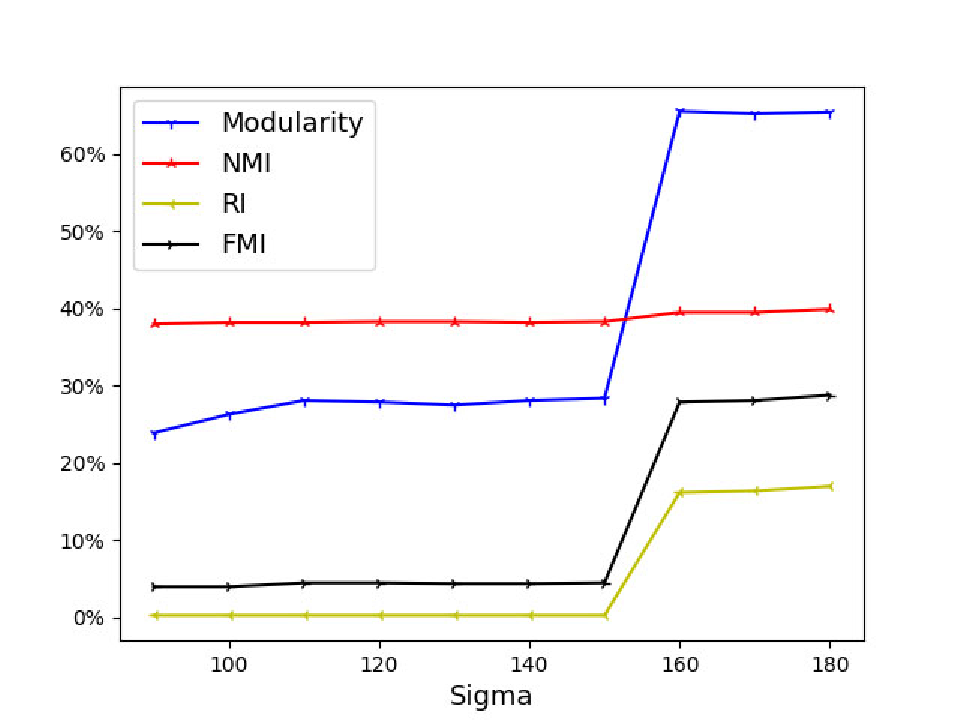}}
  \subfigure[]{
    \label{fig:5:b} 
    \includegraphics[width=2.1in,height=2.1in]{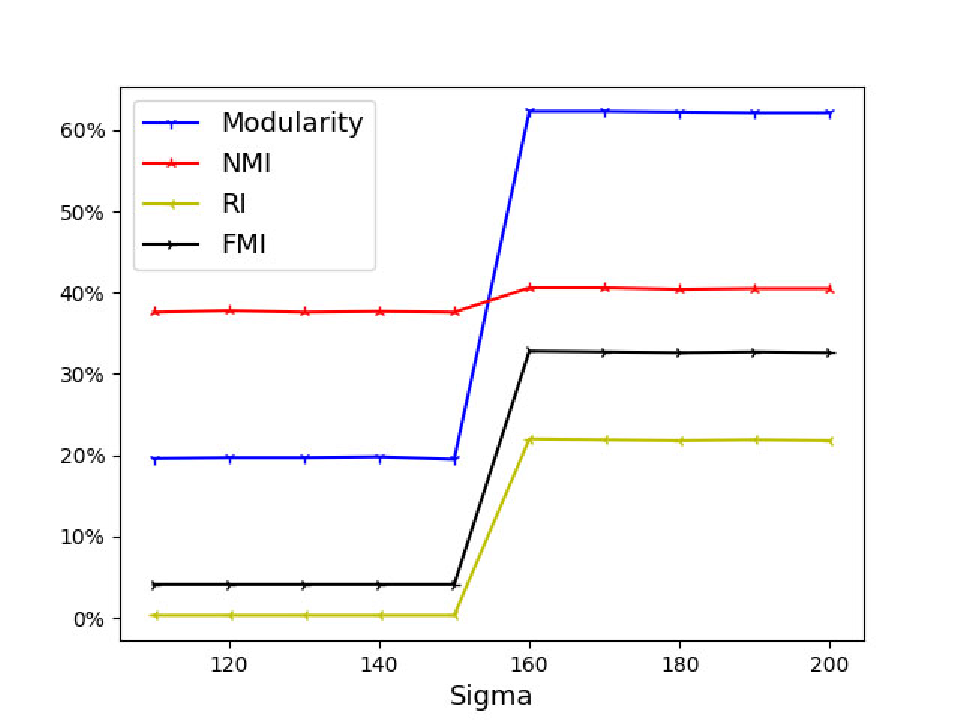}}
    \subfigure[]{
    \label{fig:5:c} 
    \includegraphics[width=2.1in,height=2.1in]{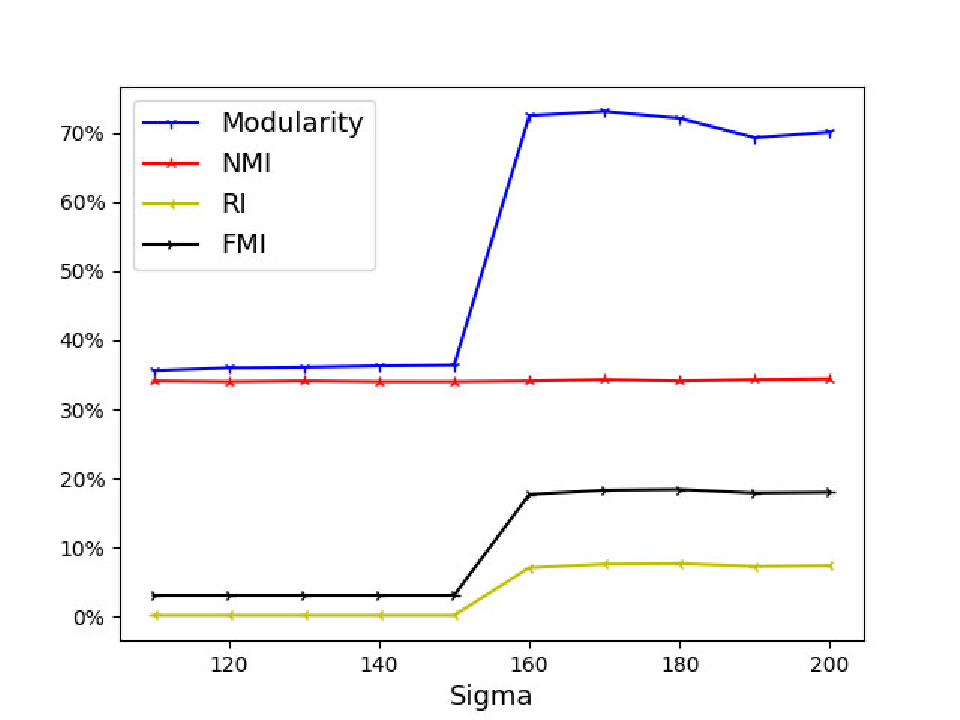}}
    \subfigure[]{
    \label{fig:5:d} 
    \includegraphics[width=2.1in,height=2.1in]{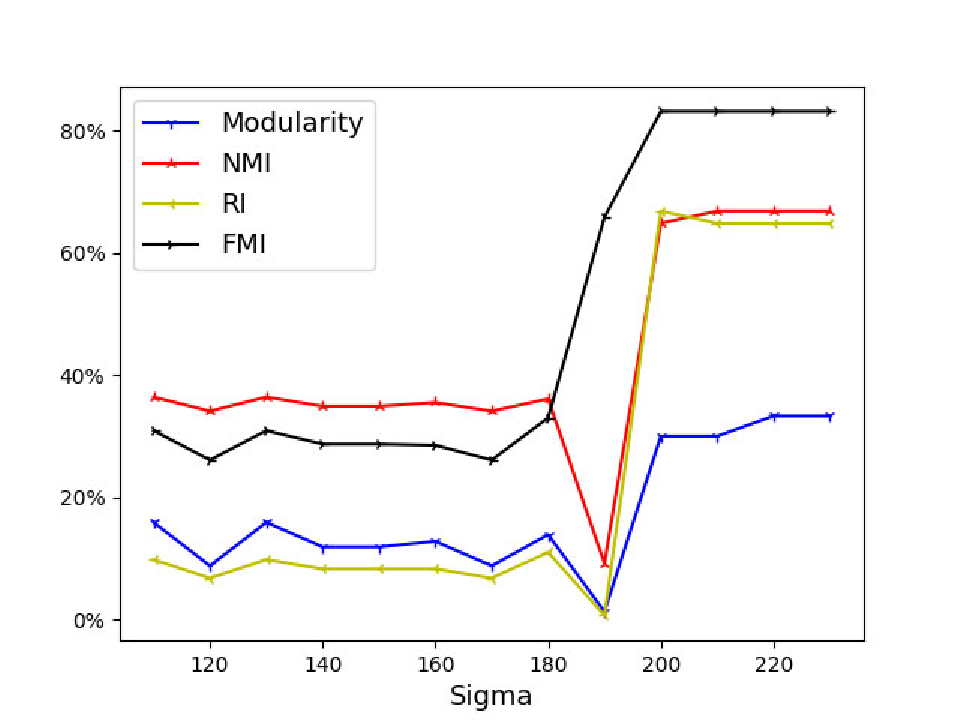}}
    \subfigure[]{
    \label{fig:5:e} 
    \includegraphics[width=2.1in,height=2.1in]{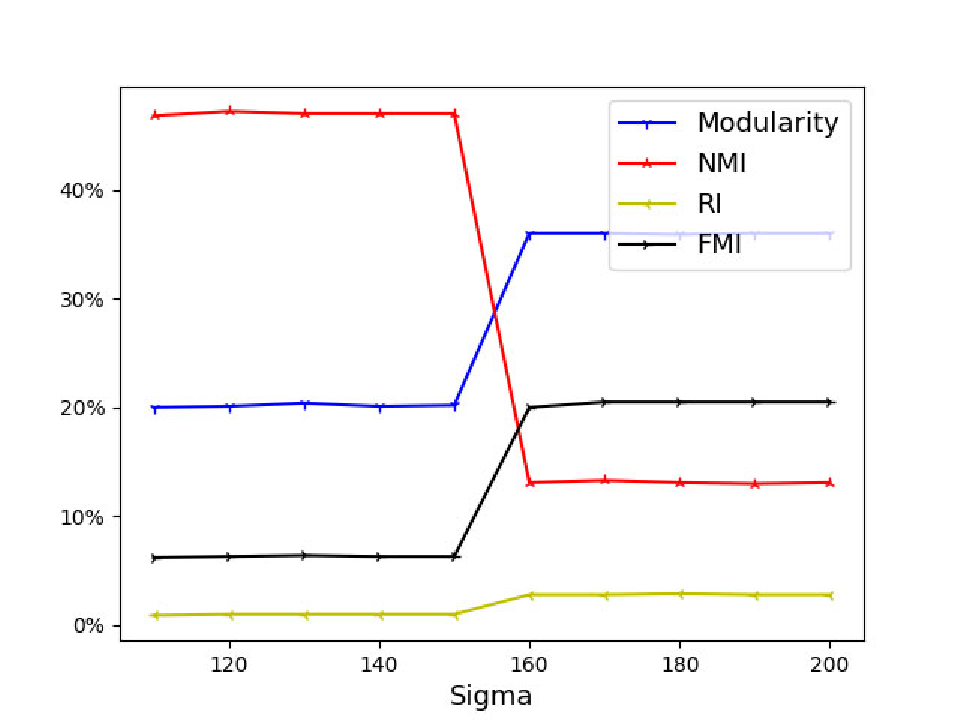}}
    
  \caption{Effect of parameter $\sigma$ on experimental results (a) Cora dataset; (b) Cora-ML dataset; (c) Citeseer dataset; (d) Karate Club datset; (e) Wiki dataset;}
  \label{fig:5} 
\end{figure}

In this paper, we extend QC to graph analysis. We develop a so-called GGD to find the minimum node of the potential function. We conduct experiments on five datasets and compare them with eight other graph clustering algorithms. The implementation of graph clustering in QC relies on $\sigma$. Below, we will provide a detailed explanation of how $\sigma$ affects the results of the algorithm. We observe the influence on the experimental results by changing the values of the parameters. According to the Fig.\ref{fig:5},  
In the five datasets used for the experiments, as the parameter $\sigma$ increase, the metrics undergo only one significant fluctuation. Before and after this fluctuation, the changes in the metrics tend to stabilize. 

Through our experiments, we have discovered that QC achieves impressive results by solely utilizing the graph structure as input, showcasing its superior flexibility compared to recent graph clustering algorithms. Additionally, by parallelizing QC, we have observed notable advantages in terms of time efficiency. Looking ahead, QC holds promise for diverse applications including bioinformatics, social network analysis, text mining, and beyond .

\section*{Acknowledgments}
Paper is supported by the Tianjin Natural Science Foundation of China (20JCYBJC00500), the Science \& Technology Development Fund of Tianjin Education Commission for Higher Education (2018KJ217). 

\section*{Declarations}

\begin{itemize}
\item Funding

This work was supported by the Tianjin Natural Science Foundation of China (20JCYBJC00500), the Science \& Technology Development Fund of Tianjin Education Commission for Higher Education (2018KJ217). 
\item Conflict of interest

No conflict of interest exists in the submission of this manuscript.
\item Ethics approval

Not applicable.
\item Consent to participate

All authors discussed the results and contributed to the writing of the paper.
\item Consent for publication

The manuscript is approved by all authors for publication.
\item Availability of data and materials

The datasets used in this work can be found in the relevant paper or on the official websites: Karate Club \url{http://vlado.fmf.uni-lj.si/pub/networks/data/Ucinet/UciData.htm}, Cora-ML, Wiki\url{https://lts2.epfl.ch/Datasets/Wikipedia/}, Cora and Citeseer \url{https://www.cs.umd.edu/~sen/lbc-proj/LBC.html}.
\item Code availability

All implementation code is visible in the \citep{Wangzhe}
\item Authors' contributions

D.L developed the idea. Z.W conducted the experiments, implemented the code and analysed the results. Z-j.H  contributed to the writing of the paper, assisted in completing the experiments and visualizing the data.
\end{itemize}


\begin{appendices}




\end{appendices}


\bibliography{ref}

\end{document}